# FULLY AUTOMATIC SEGMENTATION OF SUBLINGUAL VEINS FROM RETRAINED U-NET MODEL FOR FEW NEAR INFRARED IMAGES

*Tingxiao Yang[1], Yuichiro Yoshimura[2], Akira Morita[3], Takao Namiki[3], Toshiya Nakaguchi[2]*

[1]Graduate School of Science and Engineering, Chiba University
[2] Center for Frontier Medical Engineering, Chiba University
[3] Graduate School of Medicine, Chiba University

**ABSTRACT**

Sublingual vein is commonly used to diagnose the health status. The width of main sublingual veins gives information of the blood circulation. Therefore, it is necessary to segment the main sublingual veins from the tongue automatically. In general, the dataset in the medical field is small, which is a challenge for training the deep learning model. In order to train the model with a small data set, the proposed method for automatically segmenting the sublingual veins is to re-train U-net model with different sets of the limited number of labels for the same training images. With pre-knowledge of the segmentation, the loss of the trained model will be convergence easier. To improve the performance of the segmentation further, a novel strategy of data augmentation was utilized. The operation for masking output of the model with the input was randomly switched on or switched off in each training step. This approach will force the model to learn the contrast invariance and avoid overfitting. Images of dataset were taken with the developed device using eight near infrared LEDs. The final segmentation results were evaluated on the validation dataset by the IoU metric.

## 1. INTRODUCTION

Tongue diagnosis has extensive space for development in future home health monitoring and medical diagnosis applications. By observing the features of the tongue, it is possible to keep track the physiological function of the human body. One of the features of the tongue, which can be utilized for diagnosis, is sublingual vein. The different color and width of main sublingual veins indicate different states of the human body. To inspect the blood circulation of the body, evaluating the width of sublingual veins is required. Currently, doctors take photos of sublingual veins and use the ruler to measure the width of the veins clinically, which makes work inefficient and inaccuracy. Therefore, automatic segmentation of veins is the crucial step for sublingual veins' diagnosis.

In early studies, segmentation of sublingual veins generally needs to select two starting points of ROI (Region of Interest) [1] that destroys the automation process. In contrast, for the deep learning algorithms, once trained the model, the whole process of the segmentation will be fully automatic. In recent years, deep learning, especially Convolutional Neural Networks (CNNs) has achieved great success in computer vision. For the semantic segmentation task, the current popular CNNs are divided into two major categories. One is to extract patches from the raw image and classify the center pixel of each patch with CNN [2], [3]. The other is Fully Convolutional Network (FCN), which performs segmentation task end-to-end. By reusing shared features, the second category is much more efficient compared to the classifications of each pixel of patches. With the developing of CNNs, the state of art of the network is Mask RCNN [4], which is an extension of Faster-RCNN [5], [6]. Although, Mask RCNN has significantly outperformed the previous state of art, it is designed for multi-task, such as detection and segmentation. Generally, the model will be trained with the large dataset with many instances in each image. This is not the case in this research. The aim of this research is only to segment sublingual veins with small dataset and each image only contains limited segmentation targets. In this research, another famous CNN architecture is implemented which is so called U-Net [7]. It was designed for the pure segmentation task in medical field. To train a well-performing CNN commonly requires a large amount of data samples, which is contrary to the dataset in the medical field. The research [8] shows that to fine-tune a CNN (transfer learning) will benefit the loss convergence and performance. Therefore, the author will train U-Net model with separate steps by utilizing two sets of labels for the same training image in order to make segmentation loss easily. The first training step is to train U-Net model with tongue labels, which is much easier, compared to directly train the model with sublingual veins. The ratio of the sublingual veins of the entire image is too small to cause the loss to converge quickly.

## 2. METHOD

### 2.1. Data augmentation

Data augmentation is the conventional method for training CNNs, especially when the dataset is small, and model can easily over fit the given data. Before the data augmentation, the dataset was shuffled for each epoch. In this research, each image in the training dataset maps with 4 augmentation functions sequentially from the following list:
- Rotation with angle 10°
- Rotation with angle 20°
- Rotation with angle 30°
- Horizontal Flip

There are two more implicit data augmentation methods in this training process. One is dropout layer added after the last feature maps in the contracting part of the U-Net architecture [7]. In many researches, it has proved that by randomly shutting down the neurons before the output layer of the network will boost robustness of the performance [10]. Another implicit data augmentation is the proposed approach in which the output of segmentation will be masked on the input image (Fig. 1). By this approach, the model will learn the contrast invariance for the input image. At the same time, the number of input images for the training model is augmented.

### 2.2. Data Preprocessing

In the training process of the neural network, the initialization of the model parameters and the data distribution will affect the training efficiency at the beginning. The derivatives of the activation functions decide that the smaller values of the initial parameters and input values will speed up the learning. Moreover, the segmentation problem is always regarded as a binary classification (logistic regression) problem. The pixel values of labels are always normalized to [0, 1]. Therefore, it is necessary to normalize the input images to the same range of values.

In this research, Global Contrast Normalization (GCN) [2] was used to normalize pixel values of input image around [-1, 1] with mean zero. This processing will also enhance the contrast of input images. Every image removes the mean and is divided by the standard deviation of all the pixels of the image.

### 2.3. Batch Normalization

With deeper and deeper neural network propagation, the activation function will reduce the sensitivity of the large feature values of inputs if the weights of neurons become smaller and smaller (vanishing gradients). Alternatively, the weight of neural networks can exponentially become larger (exploding gradients). This can be derived from the general equations for the forward propagation. As known benefits for the initial normalization of input images, it is possible to normalize outputs in the hidden layer, which is called Batch Normalization (BN) [11]. BN is added after each output of convolutional layers before applying the activation function (see Fig. 1). The equations for BN is as below:

**Input:** Values of $x$ over a mini-batch:
$\mathcal{B} = \{x_i \dots x_m\}$;
Parameters to be learned: γ, β
**Output:** $\{y_i = \mathbf{BN}_{\gamma, \beta}(x_i)\}$

$$\mu_\mathcal{B} \leftarrow \frac{1}{m}\sum_{i=1}^{m} x_i \quad , \tag{1}$$

$$\sigma_\mathcal{B}^2 \leftarrow \frac{1}{m}\sum_{i=1}^{m}(x_i - \mu_\mathcal{B}) \quad , \tag{2}$$

$$\hat{x}_i \leftarrow \frac{x_i - \mu_\mathcal{B}}{\sqrt{\sigma_\mathcal{B}^2 + \epsilon}} \quad , \tag{3}$$

$$y_i \leftarrow \gamma \hat{x}_i + \beta \equiv \mathbf{BN}_{\gamma, \beta}(x_i) \quad . \tag{4}$$

BN is similar as GCN. Instead of calculating mean and the standard deviation of all pixels for each image, BN calculates those for the entire batch of data by using Eqs. (1) and (3). The final output of BN is expressed as a linear function with 2 more trainable parameters which is shown in Eq. (4). These two additional trainable parameters improve the capacity of the neural network by generating different input images for the next layer. With BN, some "dead" neurons can be reactivated or avoid the exploding gradients.

### 2.4. Regularization and Dropout

Overfitting phenomenon happens when the model has too many parameters which gives the model the ability to fit every sample in the dataset. Therefore, reducing the parameters of the model is the most efficient means of preventing overfitting.

In this research, L2 regularization is added to each convolutional layer in the network. Therefore, the final loss function form will be as below [12]:

$$\mathcal{L}_T(\mathbf{W}) = \mathcal{L}_{CE}(\mathbf{W}) + \lambda \mathcal{L}_W(\mathbf{W}) \quad , \tag{5}$$

where $\mathcal{L}_T(\mathbf{W})$ is the total loss of the network. $\mathcal{L}_{CE}(\mathbf{W})$ is the cross-entropy loss between label pixels $y^{(i)}$ and the predicted pixels $\hat{y}^{(i)}$ of final sigmoid output of segmentation. The last term is the L2 regularization loss which is the sum-of-squares of the weights. Two loss functions can be defined as the following equations:

$$\mathcal{L}_{CE}(\mathbf{W}) = \frac{1}{m}\sum_{i=1}^{m}\left(y^{(i)} \log \hat{y}^{(i)} + (1 - y^{(i)}) \log(1 - \hat{y}^{(i)})\right) \quad , \tag{6}$$

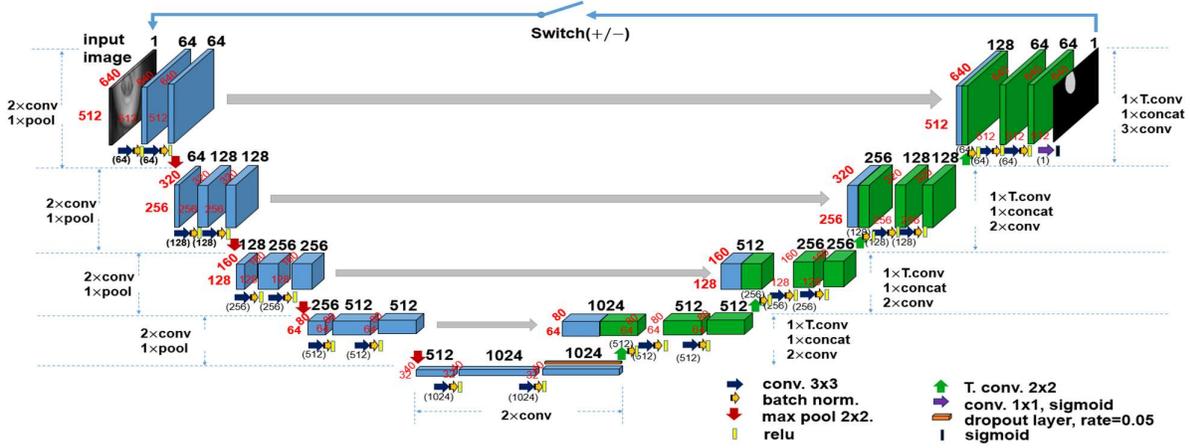

Fig. 1. Architecture of U-net. The prediction of the input image is utilized to generate the mask. The input image has 3 options: add the mask, minus the mask or keep the raw. The option can be determined by the reminder of the randomly generated integer with performing mod of 3.

$$\mathcal{L}_W(\mathbf{W}) = \frac{\lambda}{2}\|\mathbf{W}\|^2 \quad , \tag{7}$$

where $\lambda$ in Eq. (7) is the hyper-parameter which is chosen to determine the level of the penalty term of the model coefficients. This penalty term will affect the model learning rate by updating coefficients with more reduction. In this research, $\lambda$ is chosen as 0.0001 to keep the values of cross-entropy loss and L2 loss in the similar range for the initial total loss of the network.

Dropout has a similar effect to L2 regularization. With L2 regularization, some coefficients of the model will be vanished to zeros. For the same intuition as L2 regularization, dropout directly removes some neurons in the training process. Adding dropout layer at the end of last feature maps to implicitly realize data augmentation.

## 2.5. Architecture

The implemented network in this research is slightly different from the original U-Net. The detail of the entire architecture of the network is shown as Fig. 1. The only missing information in Fig. 1 is the L2 regularizer which is added in the every convolutional and transpose convolutional layer.

### 2.5.1. Statistics and Parameters

There are 23 convolutional operations in the U-Net architecture. Of these, 19 are ordinary convolutional operations with kernel size [3, 3], "same" padding and stride (1, 1). The number of filters is doubled for each convolutional step in the contract path. On the contrary, half the number of filters in expansive path; 4 of 23 are transposed convolutional operations with kernels [2, 2], stride (2, 2) which will double the dimension of the input image (upsampling). Each convolutional (including transposed convolutional) layer has a regularizer with the scale value of 0.0001 followed by BN. The output of BN continues to be the input to the activation function. Dropout is added at the end of the last feature maps with rate 0.05. In expansive path, the output of each transposed convolutional layer is concatenated with the previous feature maps which have the same dimension.

### 2.5.2. Backward Residual Augmentation

For the general residual network, the input of network block is added to the output [13]. This obviously enhances the ability of representing features, especially for the deep network. In this research, the author reversed this operation to augment the input dataset (Backward Residual Augmentation). With more and more accurate segmentation results of the training model, the model can augment input image with more specific and accurate predictions of region of interest (RoI). By masking or removing the backward residual information in the training process, the model learns the contrast invariance of RoI. Adding backward residual information with input image enhances the representation of features of RoI to make the model more sensitive to the positives in the label. Reversely, removing backward residual information makes the model more sensitive to negative samples.

## 3. EXPERIMENT

### 3.1. Dataset Preparation

The acquired infrared images were taken with dedicated device using 8 near infrared LEDs as light sources [1]. The main component of blood can absorb more near infrared than the surrounding tissue of the tongue [9]. The sublingual veins become darker than others with infrared light source. An additional component of dedicated device is the polarizing filter, which is an optical filter that passes specific polarizations of light. By tuning the polarizing filter, it is possible to filter out light-reflecting points caused by specular reflection [1].

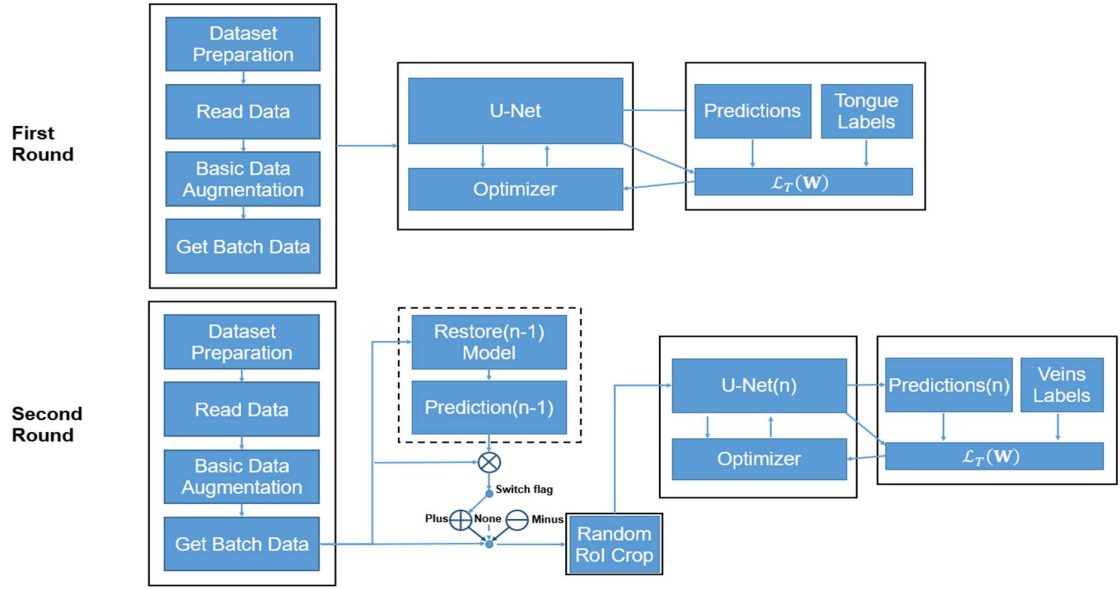

Fig. 2. The training process for the segmentation of sublingual veins. (n-1) means previous training step. (n) means the current training step. $\mathcal{L}_T(\mathbf{W})$ is the total loss for training the network, which is defined as Eq. (5). "Switch flag" is a random integer to indicate which augmentation operation to take.

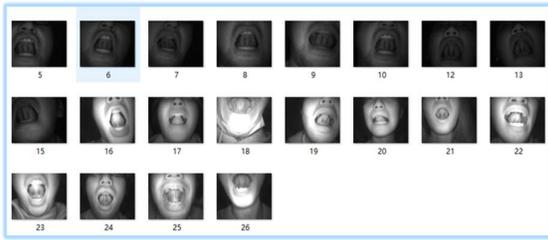

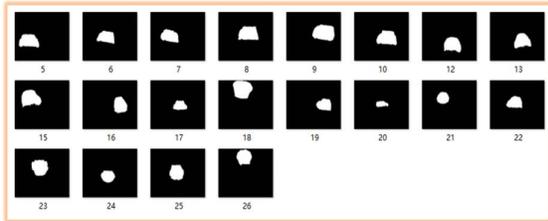

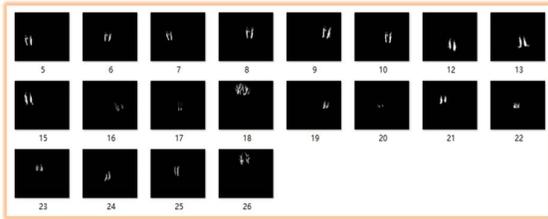

Fig. 3. The dataset for training and validation. Each raw image has two labels with respect to tongue segmentation and veins segmentation.

To make the entire training process easily repeatable for the conventional medical segmentation task, the author tries to keep the size of the training dataset to be small. In total, there are 26 images in the dataset. At first, the entire dataset is split into two parts. One is for training and validation (Fig. 3). The other is for the test. The training and validation dataset has 20 images. This dataset is further split into 5-folds. 4 images are used for validation when running the training process to decide which step to save the trained model; 16 images are for training in each fold. Finally, the different models trained by different folds were tested with the remaining 6 images. The dimension of each image in all datasets is $640 \times 512$, which has 1 channel of 8 bits value space (grayscale).

### 3.2 Training Process

The training process is shown as Fig. 2. In this research, the same U-Net model was trained by twice. Backward Residual Augmentation (BRA) is implemented in the second time training. The learning rate is 0.0001. The number of epochs in total is 100 with "early stop" strategy associated with accurate performance of validation data. The batch size is set to 2.

The first round is to train U-Net model with tongue labels. There are two reasons to pre-train the model. One is due to the ratio of the segmented target (sublingual veins) of the entire image is too small to cause the loss to converge quickly. The author has tested many times directly training the model with sublingual veins, in which it is likely that the model will be either hard to improve the IoU metric with training steps or stuck for the training loss at few initial training steps ('Direct vein Seg.' in Fig. 4, Fig. 5 and Table 1). The other reason is to speed up the experiment circle (Idea-Code-Experiment). Instead of training from the beginning, it is easier to train model with some pre-knowledge, which has the same idea as fine-tuning or transfer learning. The difference is no extra layers added to the pre-trained mode. This is the starting point of the author to get the desired performance quickly without adding extra neurons.

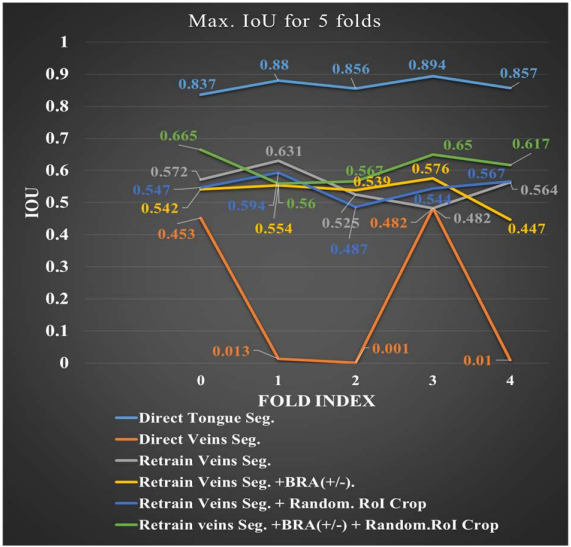
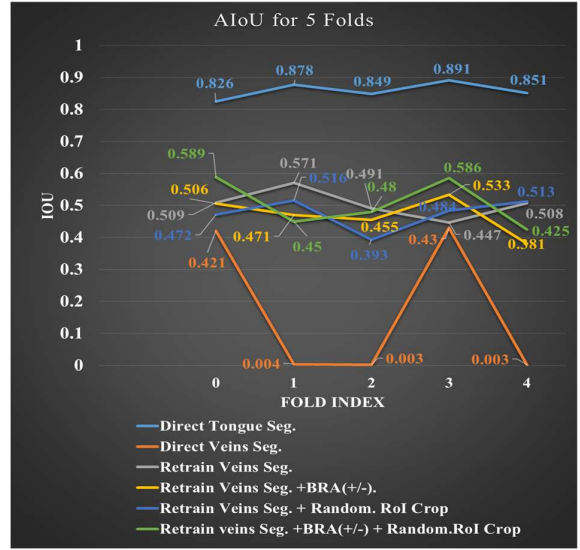

Fig. 4. Max. IoU performance for each fold tested with 6 test images. "Max. IoU" represents maximum IoU within the threshold range 0.05 to 0.95 with step 0.05 respectively.

Fig. 5. AIoU performance for each fold tested with 6 test images. "AIoU" represents average IoU within the threshold range 0.05 to 0.95 with step 0.05 respectively.

| Training Strategy | Max.IoU(at op. th.) | AIoU | Abs. Error |
|---|---|---|---|
| Direct T. Seg. | 0.864 (0.55) | 0.859 | 0.005 |
| Direct V. Seg. | 0.187 (0.80) | 0.172 | 0.015 |
| Retrain V. Seg. | 0.550 (0.80) | 0.505 | 0.045 |
| Retrain V. Seg. + BRA (+/−) | 0.521 (0.81) | 0.469 | 0.052 |
| Retrain V. Seg. + R. RoI Crop | 0.537 (0.75) | 0.476 | 0.061 |
| Retrain V. Seg. + BRA (+/−) + R. RoI Crop | **0.611** (0.85) | 0.506 | 0.105 |

Table 1. The average performance for 5 folds associated with the results in both Fig. 4 and Fig. 5. "Abs. Error" is the absolute error between Max. IoU and AIoU. "R. RoI Crop" means random RoI crop which has been explained in the training process section. "T. Seg." means tongue segmentation. "V. Seg." means sublingual vein segmentation.

In the second training round, firstly to use previous trained model from the first round to predict the tongue labels. Multiply this tongue segmentation with the raw input image to get the mask of RoI. By randomly generating the integers which calculates the mod of 3. There are 3 options for the reminder (0, 1, 2). When the reminder is 0, add the mask to the input image. 1 indicate that the input image minus the mask. 0 means to send the input image directly into the model without any BRA. As mentioned in the previous section, the idea behind these 3 options is to make the model learning the contrast invariance of RoI.

After "switch", the processed image will be randomly cropped according to the input label. The minimum cropped bounding box is defined as [ymin, xmin, ymax, xmax], where 4 values are determined by the input label. The left top corner of the bounding box will be randomly changed from (0, 0) to (ymin, xmin); the right bottom corner varied from (ymax, xmax) to (1, 1). At the same time, to keep the capability for segmenting the input images without any pre-augmentations for the future validation and test. No BRA and random crop are applied on validation and test images. All the trained models for different folds (mentioned in 3.1) will be tested by 6 images. The related results will be shown in the Results chapter.

### 3.3. Intersection over Union

The final segmentation results are evaluated by the Intersection over Union (IoU), which is also named as Jaccard Index defined as follows:

$$IoU = \frac{|A \cap B|}{|A \cup B|} \quad . \tag{8}$$

The segmentation result can also be regarded as the binary classification result. In this case, Eq. (8) can be calculated from the confusion matrix as interpreted as:

$$\text{IoU} = \frac{tp}{tp + fp + fn} \quad , \quad (9)$$

where $tp$ means the number of true positives for the binary classification (segmentation result). $fp$ is the number of false positives; $fn$ is the number of false negatives. Hence, Eq. (9) can be interpreted as the intersection of true positives over the union of all segmented positives.

## 4. RESULTS

The number of images used to train the model is 16. The average performance of 5 folds for directly training the raw U-Net model to segment tongue and sublingual veins is shown in Table 1. From Table. 1, The results were tested with additional test dataset for both tongue and sublingual veins segmentation. "Max. IoU" and "AIoU" represent the maximum IoU at optimal threshold and average IoU values, respectively, within the threshold range [0.05, 0.95] with step 0.05. "Direct" means end-to-end training without pre-training the model which means pure U-Net segmentation process. "Retrain" means firstly the model was trained for tongue segmentation and restore the saved tongue segmentation model to train for sublingual veins segmentation. IoU values are calculated according to Eq. (9).

The Table 1 also shows that the raw U-Net architecture is more sensitive to the large object (in tongue segmentation case). In the other words, U-Net doesn't work well when the ratio of segmented target for each entire image is very small (vein segmentation case). Both maximum IoU and average IoU performance are lower than 0.2 for direct training the network to segment the sublingual veins. From Fig. 4, it shows that individually applying BRA and random Crop in the training process will not directly benefit the performance too much. But combine the BRA and random RoI crop which will significantly improve the segmentation results. The average IoU (AIoU) within the entire threshold span will indicate the binary classifier performance for both positives and negatives which is similar as AUC (Area Under Curve) for the IoU metric. Moreover, the "Max. IoU" metric shows the best performance at the optimum threshold. In Fig. 6, it shows 4 kinds of plots. The column (a) is the raw test input image; the column (b) is the true labels. The columns (c) and (d) are the original segmentation result and the optimized segmentation result, respectively. The results in Fig. 6 were generated by the model 0 (fold 0) with the training strategy "Retrain V. Seg. + BRA (+/−) + Random. RoI Crop" as shown in Fig. 4 with a maximum IoU performance 0.665. Although, the maximum IoU performance is very good, the absolute error between the Max. IoU and AIoU reaches to 0.105 which is the largest in the Table 1. This indicates more false positives are generated, since the optimum threshold is at 0.85. This means the model needs higher threshold to suppress the false positives.

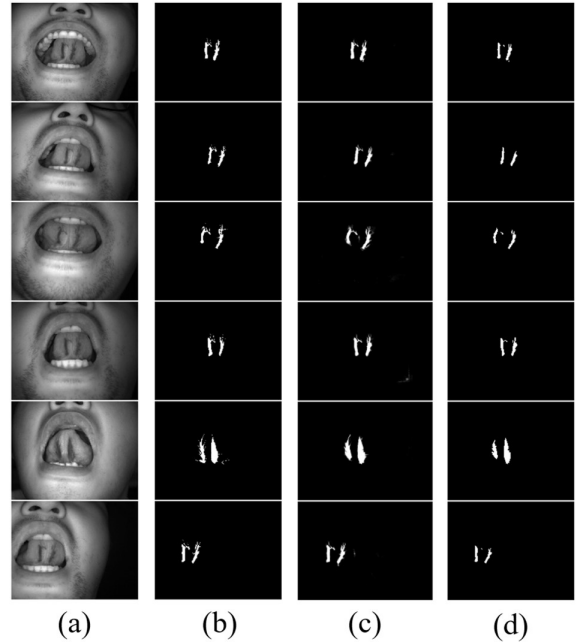

Fig. 6. Segmentation Results. (a) raw input images; (b) raw true labels; (c) raw output of segmentation; (d) optimized segmentation results at threshold 0.85.

## 5. CONCLUSION

This study applies a convolutional neural network to segment the sublingual veins. From the showed results, the models trained in this research have the significant performance. However, the performance needs to be better to apply to practical application (consider false positives). In the future research, to make the model learn richer information of features is the first thing to consider. "Group convolution" and "dense connection" will be implemented in the future study to improve the performance without adding extra neurons. "Pyramid Feature" or "Pyramid receptive field" can also be a good choice to improve the performance. In this research, the training process is not end-to-end (two round training process). Hence, end-to-end training or joint training will also be considered in the future study.